\documentclass[10pt, journal]{IEEEtran} 

\usepackage{graphicx} 
\usepackage{amsmath} 
\usepackage{float}          
\usepackage{algorithm}       
\usepackage{algpseudocode}   
\usepackage{amsfonts}  
\usepackage{stfloats}
\usepackage{booktabs} 
\usepackage{array}   
\usepackage{diagbox}
\usepackage{hyperref}
\usepackage{balance}
\begin{document}

\title{Iterative Diffusion-Refined Neural Attenuation Fields for Multi-Source Stationary CT Reconstruction: NAF Meets Diffusion Model}
\author{Jiancheng Fang, Shaoyu Wang, Junlin Wang,  Weiwen Wu, \IEEEmembership{Member, IEEE}, \\Yikun Zhang, Qiegen Liu, \IEEEmembership{Senior Member, IEEE}
\thanks{This work was supported by the National Natural Science Foundation of China (U24A20304), and in part by the National Key Research and Development Program of China (Grant 2023YFF1204300 and Grant 2023YFF1204302). (Jiancheng Fang and Shaoyu Wang are co-first authors.) (Corresponding author: Qiegen Liu).}
\thanks{Qiegen Liu, Jiancheng Fang and Shaoyu Wang are with the School of Information Engineering, Nanchang University, Nanchang 330031, China; and Junlin Wang is with the School of Mathematics and Computer Sciences, Nanchang University, Nanchang 330031, China (e-mails: liuqiegen@ncu.edu.cn; d5283377@163.com; wangshaoyu@ncu.edu.cn; 2225046979@qq.com).}
\thanks{Weiwen Wu is with the School of Biomedical Engineering, Sun Yat-sen University, Shenzhen 518000, Guangdong, China (e-mail: wuweiw7@mail.sysu.edu.cn).}
\thanks{Yikun Zhang is with the Laboratory of Image Science and Technology, School of Computer
Science and Engineering, and the Key Laboratory of New Generation Artificial Intelligence Technology and Its Interdisciplinary Applications, Ministry of Education, Southeast University, Nanjing 210096, China(e-mail: yikun@seu.edu.cn).}
}
\date{}

\markboth{Journal of 	IEEE Transactions on Computational Imaging}%
{Shell \MakeLowercase{\textit{et al.}}: A Sample Article Using IEEEtran.cls for IEEE Journals}

\maketitle

\begin{abstract}
Multi-source stationary computed tomography (CT) has recently attracted attention for its ability to achieve rapid image reconstruction, making it suitable for time-sensitive clinical and industrial applications. However, practical systems are often constrained by ultra-sparse-view sampling, which significantly degrades reconstruction quality. Traditional methods struggle under ultra-sparse-view settings, where interpolation becomes inaccurate and the resulting reconstructions are unsatisfactory. To address this challenge, this study proposes Diffusion-Refined Neural Attenuation Fields (Diff-NAF), an iterative framework tailored for multi-source stationary CT under ultra-sparse-view conditions. Diff-NAF combines a Neural Attenuation Field representation with a dual-branch conditional diffusion model. The process begins by training an initial NAF using ultra-sparse-view projections. New projections are then generated through an Angle-Prior Guided Projection Synthesis strategy that exploits inter view priors, and are subsequently refined by a Diffusion-driven Reuse Projection Refinement Module. The refined projections are incorporated as pseudo-labels into the training set for the next iteration. Through iterative refinement, Diff-NAF progressively enhances projection completeness and reconstruction fidelity under ultra-sparse-view conditions, ultimately yielding high-quality CT reconstructions. Experimental results on multiple simulated 3D CT volumes and real projection data demonstrate that Diff-NAF achieves the best performance under ultra-sparse-view conditions.
\end{abstract}


\section{INTRODUCTION}\label{sec:intro}
Cone-Beam Computed Tomography (CBCT) is a non-destructive imaging technique that reconstructs 3D internal structures from multi-view X-ray projections based on ray penetration~\cite{3}. Conventional rotating-source CBCT requires long scan times and delivers a high radiation dose, which raises cost and limits use. In recent years, multi-source stationary CT has attracted wide attention because it uses multiple fixed X-ray sources and static detectors, provides high temporal resolution with inherently lower dose, and enables ultra-high-speed, high-resolution tomography~\cite{stationCT2}. It has been applied effectively to dynamic imaging, industrial inspection, and cardiac motion, among other applications~\cite{67}. However, due to practical system constraints, reconstruction is typically sparse-view because limited angular sampling violates the Nyquist criterion, which makes the inverse problem ill-posed and prone to artifacts~\cite{8,9}.
\begin{figure}[t]
    \centering
    \includegraphics[width=\columnwidth]{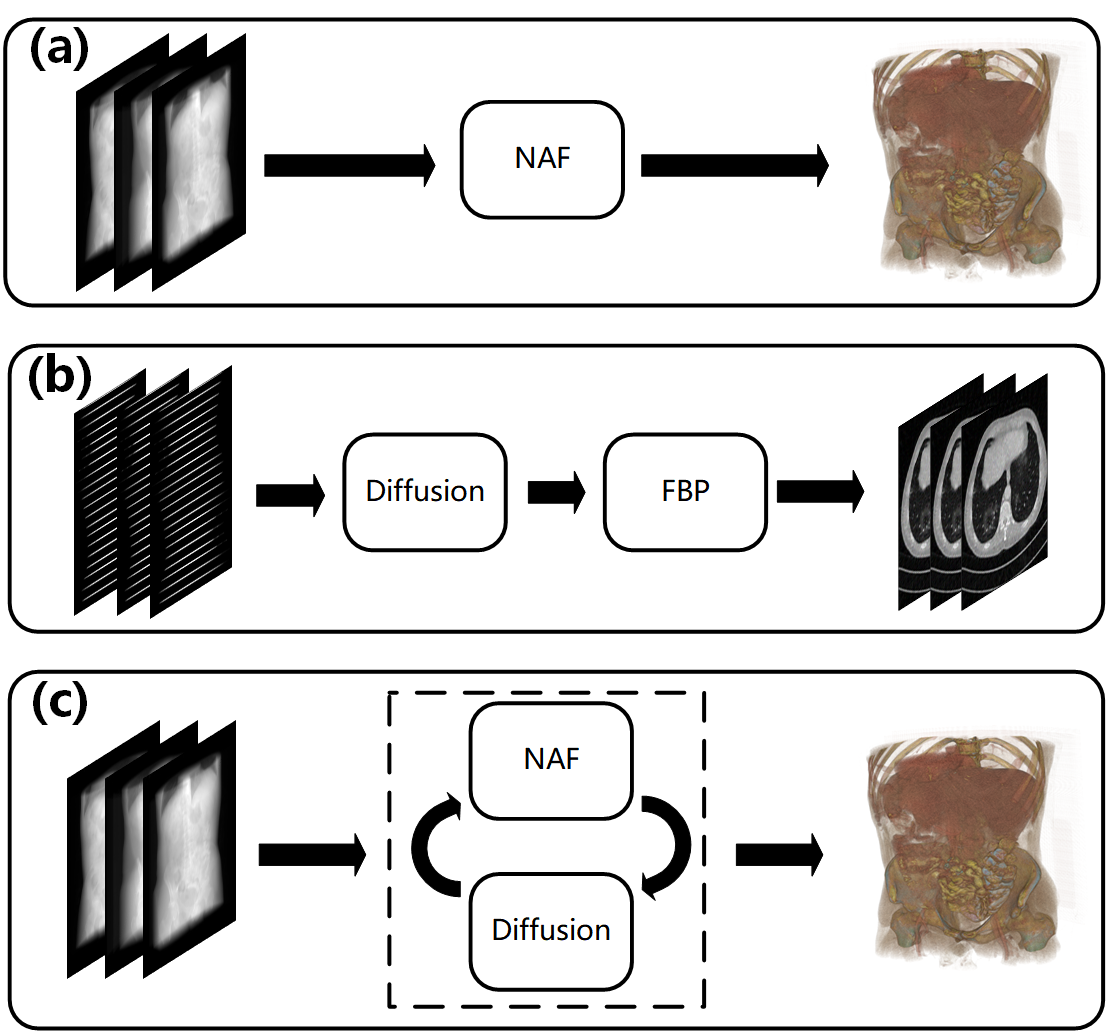}
    \caption{Overview of the reconstruction tasks and our method. (a) The process of performing 3D reconstruction from DR projections using NAF. (b) The process of reconstructing image slices from sinograms using the diffusion model. (c) Our proposed framework: DR projections are used as input, followed by an iterative process between NAF and the diffusion method, ultimately producing a high-quality 3D reconstruction.}
    \label{fig:method_overview}
\end{figure}
Early CT reconstruction methods include both analytical and iterative approaches. Analytical methods, such as filtered backprojection (FBP)~\cite{FBP},  are highly susceptible to pronounced streak artifacts under sparse-view sampling. Iterative methods exemplified by the Simultaneous Algebraic Reconstruction Technique (SART)~\cite{SART} suffer from slow computation due to a large number of iterations and still fail to fully eliminate the streak artifacts present in analytical methods. Building on iterative reconstruction, some studies have further introduced compressed sensing (CS) approaches, including total variation (TV)~\cite{12}, wavelet transforms~\cite{WaveletRegularization}, dictionary learning~\cite{DictionaryLearning}, and $L_0$ regularization~\cite{L0}. However, these methods require manual prior design and extensive parameter tuning. Moreover, since they are also iteration-based, their computational cost remains high. With the rapid development of deep learning, deep learning methods have been widely applied to CT reconstruction. These methods can be categorized into data-driven and data–model-driven approaches. Data-driven approaches demand large, carefully paired training datasets and lack explicit physical constraints, which reduces interpretability~\cite{End2endprojection1,End2endprojection2,7,13,15,18,20}. In contrast, data–model-driven approaches, while incorporating physical and geometric constraints~\cite{Celldatamodeldrive,17,19}, remain iteration-based and require significant computational resources.

Compared with traditional deep learning models, implicit neural representations (INRs) encode signals as continuous functions parameterized by neural networks, mapping coordinates to values for flexible and resolution-independent modeling~\cite{30}. For CT reconstruction, neural fields represent the volumetric X-ray attenuation map as a continuous function and are optimized with a differentiable forward projector so that simulated projections match measured data~\cite{14,31}. Building on this formulation, recent studies introduce self-supervised geometric consistency, motion-aware formulations, efficient encodings, and specialized parameterizations such as the Neural Attenuation Field (NAF). Representative examples include InTomo~\cite{32}, neural-field-based CT reconstructions~\cite{33}, and NAF~\cite{34}. As shown in Fig.~\ref{fig:method_overview}(a), a typical NAF-based approach consumes multi-view digital radiography (DR) projections and directly reconstructs a 3D volume, which provides strong inter-slice continuity. Under ultra-sparse-view acquisition, however, it often relies on inaccurate interpolation and the reconstructions become blurred.

Diffusion models are another class of generative models. They learn complex data distributions by gradually adding noise and then learning to reverse this process, which turns the learned distribution into effective priors~\cite{23,24,25,26}. These models have emerged as an effective approach for CT reconstruction. Our previous work has focused on CT reconstruction using diffusion models in the raw-data domain~\cite{SWORD,29}. As illustrated in Fig.~\ref{fig:method_overview}(b), a simplified algorithmic workflow first applies a diffusion model to the sinogram projections, for example to complete missing views or to denoise them, and then backprojects the processed data to obtain the CT reconstruction. This progressive, distribution-based learning supports unsupervised training, strong generalization, and high-quality reconstructions. However, directly applying diffusion models to full 3D reconstruction often incurs substantial computational cost and leads to slow imaging.

Building on the above, we believe that combining diffusion models with NAF is highly promising. The fast 3D imaging and stable inter-slice continuity offered by NAF are expected to complement the high-quality, high-resolution reconstructions enabled by diffusion models, which motivates the joint design explored in this work. Previous studies have explored preliminary attempts to combine these two approaches and have achieved certain pioneering results~\cite{implicitdiffusion1,implicitdiffusion2}. However, these methods still fail to address the 3D reconstruction task of multi-source stationary CT under ultra-sparse-view conditions. To overcome this limitation, this study proposes Diff-NAF. As shown in Fig.~\ref{fig:method_overview}(c), the proposed method tightly integrates diffusion models with NAF and establishes a new iterative CT reconstruction framework. Our approach makes three noteworthy contributions as follows:

\begin{itemize}
  \item This paper proposes Diffusion-Refined Neural Attenuation Fields (Diff-NAF), an iterative optimization framework focused on the DR-domain and designed for fast imaging in multi-source stationary CT under ultra-sparse-view conditions. To the best of our knowledge, this is the first algorithm to achieve deep coupling between NAF and diffusion models in the DR-domain.
  \item We propose the Diffusion-driven Reuse Projection Refinement (DRPR) module, which implements a DR projection refinement process through dual-branch conditional diffusion and a Dynamic Range Adaptive Transformation (DRAT). This design addresses the lack of physical consistency constraints in DR projections and resolves the consistency issues caused by projection reuse.
  \item Our approach introduces the Angle-Prior Guided Projection Synthesis (APGPS) strategy to guide the generation of new view projections and improve training efficiency.
\end{itemize}

Relevant background on the imaging principles of multi-source stationary CT and on the CT reconstruction principles of diffusion models and NAF is reviewed in Section~\ref{sec:preliminary}. The detailed algorithmic workflow of the proposed method is presented in Section~\ref{sec:method}. Section~\ref{sec:experiments} provides the dataset description, experimental setup, and analysis of results. Section~\ref{sec:conclusion} concludes with a summary and discussion.

\section{PRELIMINARY}\label{sec:preliminary}
\subsection{Overview of a Multi-Source Stationary CT System}
The hardware design of CT systems must consider the choice of X-ray sources, the scanning mechanism, and the detector. These factors directly influence imaging speed, stability, and reconstruction quality. Rotating scanners with one or two thermionic tubes remain mainstream~\cite{Stationary_new1}. With a single source–detector pair, systems are typically circular or helical CT scanners. The source–detector assembly rotates mechanically to collect projections from multiple views. The technology is mature and the cost is moderate, so such scanners are widely used in clinical and industrial settings. However, mechanical rotation limits scanning speed and can introduce motion artifacts, particularly in dynamic imaging.

\begin{figure}[htbp]
\centering
\includegraphics[width=\columnwidth]{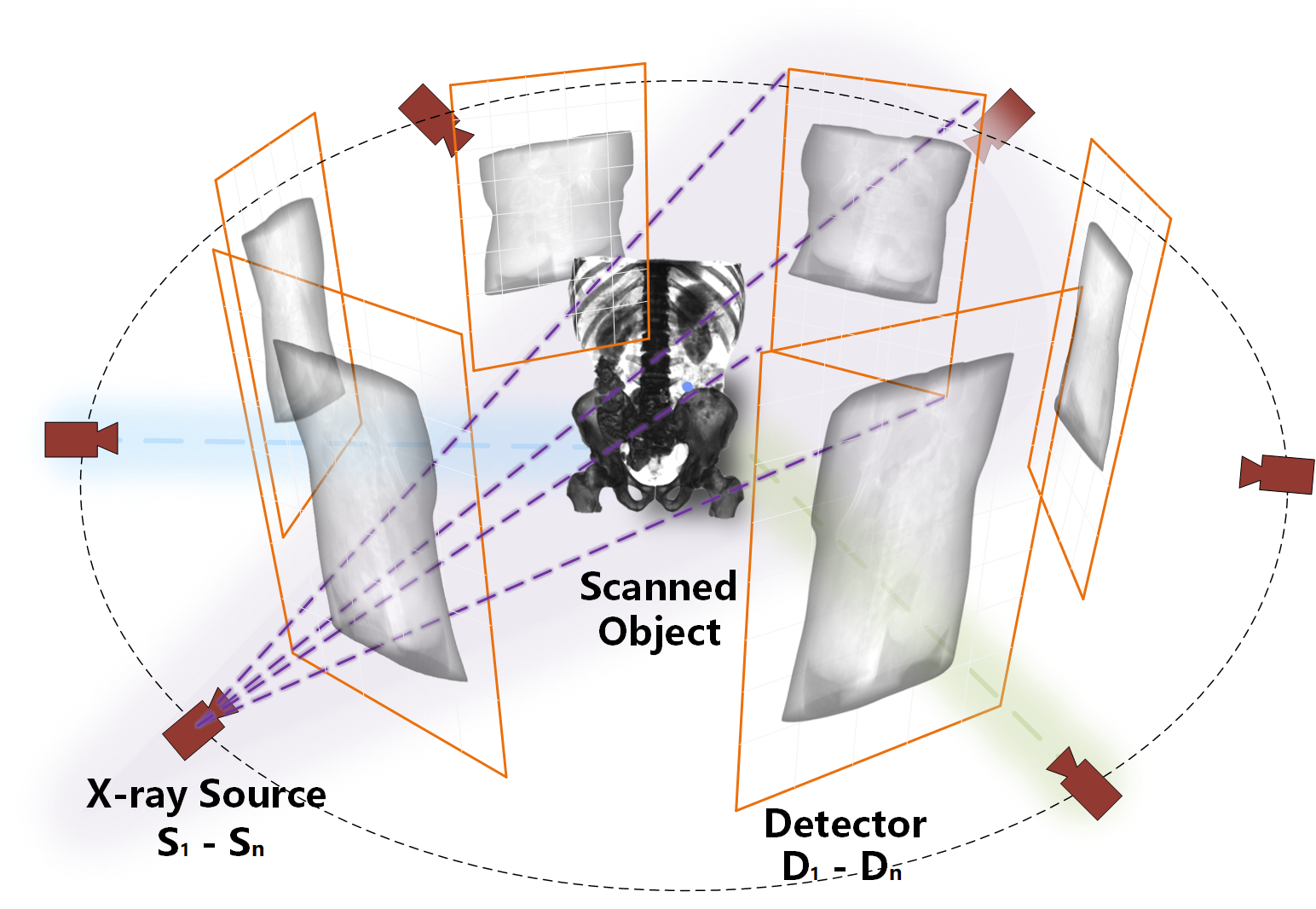}
\caption{Schematic of a multi-source stationary CT configuration. Multiple fixed X-ray sources ($S_1$–$S_n$) and corresponding detectors ($D_1$–$D_n$) are arranged around the object, where $n$ is typically less than 30. During acquisition, the sources are activated sequentially or in a time-multiplexed sequence, and the detectors record DR projections from different directions.}
\label{fig:multi_source_CT}
\end{figure}

As an alternative, multi-source stationary CT has advanced rapidly and is well suited for fast and dynamic imaging. The gantry does not rotate. Multiple fixed source–detector pairs are arranged around the object, and each source is triggered electronically, either sequentially or in parallel~\cite{Stationary_new1,67}. Fig.~\ref{fig:multi_source_CT} illustrates a representative multi-source stationary CT configuration. To support fast switching, stationary systems often use cold-cathode X-ray sources instead of thermionic tubes. This design enables precise timing and rapid acquisition for high temporal resolution 3D imaging.

During imaging, X-rays attenuate as they traverse the object, and the detectors measure the transmitted intensity to form DR projections. The process follows the generalized Beer–Lambert law given by
\begin{equation}
I = \int I_0(E)\,\exp\!\left(-\int \mu(l,E)\,\mathrm{d}l\right)\,\mathrm{d}E,
\end{equation}
where $I_0(E)$ denotes the incident X-ray spectrum at energy $E$, $\mu(l,E)$ is the energy-dependent linear attenuation coefficient along ray path $l$, and $I$ is the transmitted intensity.

Although multi-source stationary CT offers clear advantages for fast 3D imaging, practical constraints such as hardware cost and geometric layout often limit the number of source–detector pairs. As a result, only a small number of projection views can be acquired, leading to an ultra-sparse-view reconstruction problem. Traditional FDK methods produce severe sparse-view artifacts~\cite{FBP}. Iterative and compressed-sensing approaches require long reconstruction times, limiting practical deployment~\cite{ART,SART}. Deep learning-based methods typically demand large paired datasets~\cite{End2endprojection1,End2endprojection2}. Therefore, there is an urgent need for algorithms specifically tailored to multi-source stationary CT that enable fast and high-quality 3D reconstruction under ultra-sparse-view conditions.

\subsection{Diffusion Models for CT Reconstruction}
Recent studies have demonstrated the effectiveness of diffusion models for generation and denoising tasks~\cite{23,59}. Ho et al. introduced Denoising Diffusion Probabilistic Models (DDPM), which formulate generation as a discrete-time Markov process consisting of a forward noising procedure and a reverse denoising process~\cite{42}. A time-conditioned neural network is trained to iteratively recover the data. Subsequently, Song et al. generalized this framework to continuous time using stochastic differential equations (SDEs)~\cite{68}. By learning the score function, samples can be generated by integrating the reverse-time SDE, thereby unifying diffusion models with score-based generative modeling. Taking DDPM as an example, the forward and reverse processes are given as follows:
\begin{equation}
q(\mathbf{x}_t \mid \mathbf{x}_{t-1})
= \mathcal{N}\!\left(\mathbf{x}_t;\, \sqrt{1-\beta_t}\,\mathbf{x}_{t-1},\, \beta_t\mathbf{I}\right),
\label{eq:6}
\end{equation}
\begin{equation}
p_\theta(\mathbf{x}_{t-1} \mid \mathbf{x}_t)
= \mathcal{N}\!\left(\mathbf{x}_{t-1};\, \mu_\theta(\mathbf{x}_t, t),\, \tilde{\beta}_t \mathbf{I}\right).
\label{eq:7}
\end{equation}

Here, $\beta_t$ is a predefined noise schedule, $\mu_\theta(\mathbf{x}_t, t)$ is parameterized by the denoising network, and $\tilde{\beta}_t$ denotes the variance term derived from the forward process.

For sparse-view CT imaging, prior work falls into three categories: sinogram-domain methods, image-domain methods, and dual-domain methods. Sinogram-domain methods align with the imaging physics, impose explicit constraints, and yield stable reconstructions. Examples include GMSD, which trains a score-based generative model on full-view sinograms with a multi-channel strategy~\cite{47}, and SWORD, which performs sub-band diffusion on wavelet-decomposed sinograms and uses a unified optimization with wavelet sparsity to stabilize training~\cite{SWORD}. Yet these methods are sensitive to noise and missing views. Image-domain methods directly produce reconstructions with good visual quality. CT-SDM improves the degradation operator and the diffusion sampling mechanism~\cite{46}, and Chung et al.\ combine a pre-trained 2D diffusion prior with test-time consistency along the third dimension for volumetric recovery~\cite{24}. Nevertheless, these methods face under-determinacy, and artifacts and local minima occur. Dual-domain methods combine the strengths of both. For example, Liu et al.\ inject diffusion priors in both sinogram and image domains with a dynamic temporal weighting mask to suppress metal artifacts~\cite{27}. However, these methods incur higher runtime.

The application of diffusion models in the DR-domain remains underexplored. Although some studies modify the noise model to denoise DR projections~\cite{48}, these methods do not address ultra-sparse-view CT reconstruction. Given the strong generalization ability of diffusion models and the importance of DR projections for 3D reconstruction, a new diffusion-model approach that operates directly in the DR-domain is urgently needed.

\subsection{Neural Attenuation Fields}
INRs are a type of method based on coordinate-conditioned neural networks that model a continuous field. In recent years, 3D reconstruction methods have increasingly adopted implicit representations~\cite{49,50}. In the field of scene rendering, Mildenhall first proposed Neural Radiance Fields (NeRF). The framework pioneered representing scenes as a continuous 5D function for novel view synthesis~\cite{30}. Inspired by this work, Zha et al. proposed the NAF framework for CBCT reconstruction by modeling attenuation coefficients while preserving the linear imaging principle, achieving fast CBCT imaging~\cite{34}.

Recent studies have advanced NAF in various directions. NeAT introduces an adaptive hierarchical neural rendering framework that combines neural features with adaptive explicit representations to balance computational efficiency and reconstruction quality in multi-view inverse rendering~\cite{51}. The INeAT framework employs an iterative pose optimization strategy with a feedback mechanism to refine input image poses, reducing reliance on prolonged CT scanning and specialized hardware~\cite{52}. Fang et al. proposed a novel view enhancement strategy for neural attenuation fields, mitigating detail loss in sparse-view CT reconstruction due to insufficient data~\cite{53}. Liu proposed a geometry-aware encoder–decoder framework that maintains geometric consistency between 3D images and 2D projections during backprojection, demonstrating high reconstruction quality and efficiency on both simulated and real datasets~\cite{55}. The SAX-NeRF method recently introduced a Lineformer architecture with an MLG ray sampling strategy, achieving strong performance in novel view synthesis~\cite{54}.

Current ultra-sparse-view CT reconstruction methods based on NAF show advantages in suppressing streak artifacts. However, their reconstruction results still have limitations. The main problems include insufficient spatial resolution and noise interference in dark cavity regions.

\section{METHOD}\label{sec:method}
\subsection{Motivation}

\begin{figure}[t]
    \centering
    \includegraphics[width=\columnwidth]{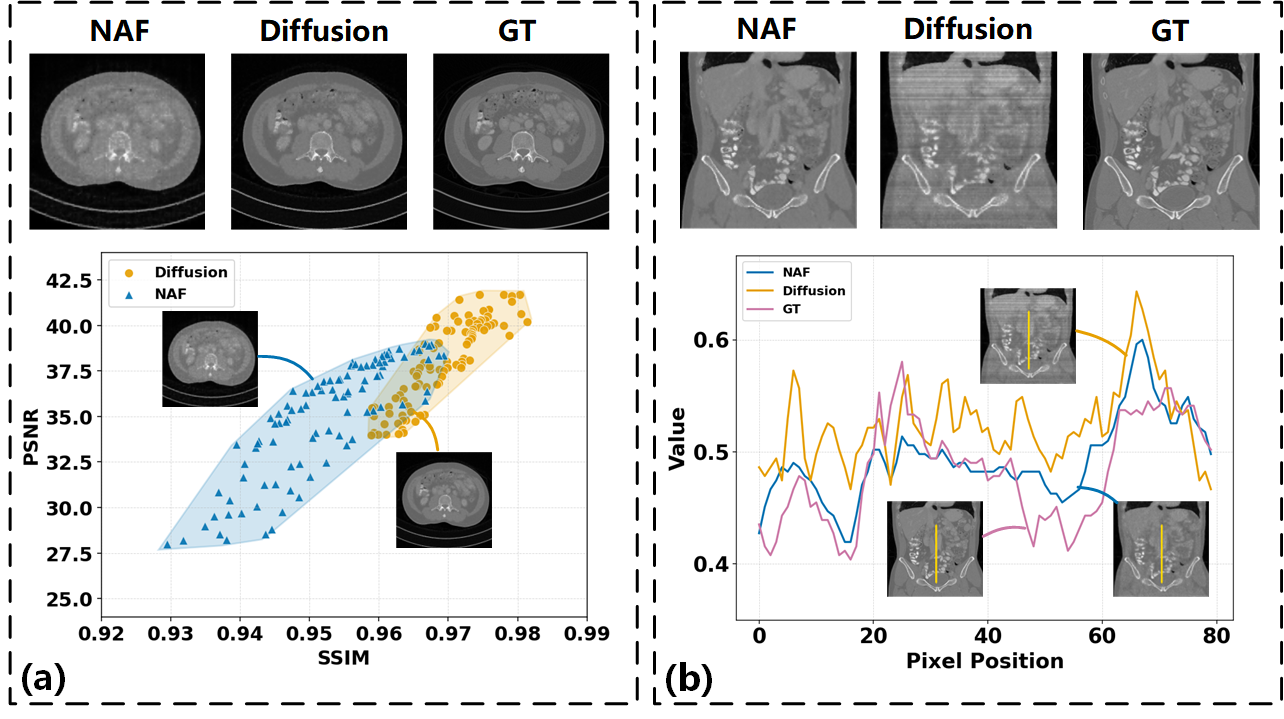}
    \caption{Motivation schematic. We consider a setting where NAF is trained on full 3D volumes, whereas the diffusion model is trained on axial slices. (a) On axial slices, the diffusion model delivers higher quality and greater stability than NAF. (b) On the coronal plane, NAF still performs well; however, the diffusion approach—constructed by stacking axial predictions and re-slicing along the coronal axis—exhibits severe inter-slice inconsistencies.} 
    \label{fig:new_4} 
\end{figure}

Multi-source stationary CT systems use electronic control to coordinate the emission of multiple X-ray sources and can complete CBCT scans in a very short time. However, due to hardware costs and geometric constraints, most practical systems only have a limited number of source–detector pairs. This leads to an ultra-sparse-view sampling scenario, where the reconstructed images often suffer from prominent streak artifacts and the loss of high-frequency details.

NAF models the 3D attenuation field with a coordinate-based implicit representation and is optimized directly from sparse-view projections, enabling fast volumetric reconstruction without large external datasets. Under ultra-sparse-view, however, under-constrained regions lean on interpolation, blurring true boundaries and suppressing high-frequency details. With few projections, optimization drifts into large suboptimal basins, yielding over-smoothed, coarse results. This is evident in Fig.~\ref{fig:new_4}(a): given the same number of views, NAF outputs a full 3D volume, yet its per-slice quality is generally inferior to diffusion-based reconstructions.

Diffusion models provide strong structural and textural priors and can produce high-quality 2D slices along the training-aligned direction even under extreme sparsity. Their slice-wise refinement, however, breaks volumetric coherence and induces inter-slice inconsistency along the orthogonal direction, as shown in Fig.~\ref{fig:new_4}(b). Building on the above analysis, we contend that NAF and diffusion models are highly complementary and have strong potential for integration. However, prior attempts to combine diffusion priors with NAF~\cite{implicitdiffusion1,implicitdiffusion2} remain insufficient in ultra-sparse-view regimes.

To address this, we propose Diff-NAF, a framework that deeply couples diffusion models and NAF. The framework first uses NAF to quickly produce an initial coarse reconstruction constrained by the limited real projections. It then applies forward projection to generate supplementary virtual projections from additional views. A dual-branch conditional diffusion model refines these DR projections, which lack strict physical constraints, and enhances their detail and structural fidelity. The refined projections are then fed back to guide the next round of NAF optimization. By repeating this cycle, the reconstruction quality gradually improves. Throughout the iterative process, the newly introduced projections act as pseudo-annotations and provide additional information that drives NAF optimization. As a result, NAF is able to escape its initial local optimum and gradually approach a more global optimum. 

\subsection{Iterative Framework}
\begin{figure*}[t]
    \centering
    \includegraphics[width=\textwidth, height=0.5\paperheight, keepaspectratio, scale=0.9]{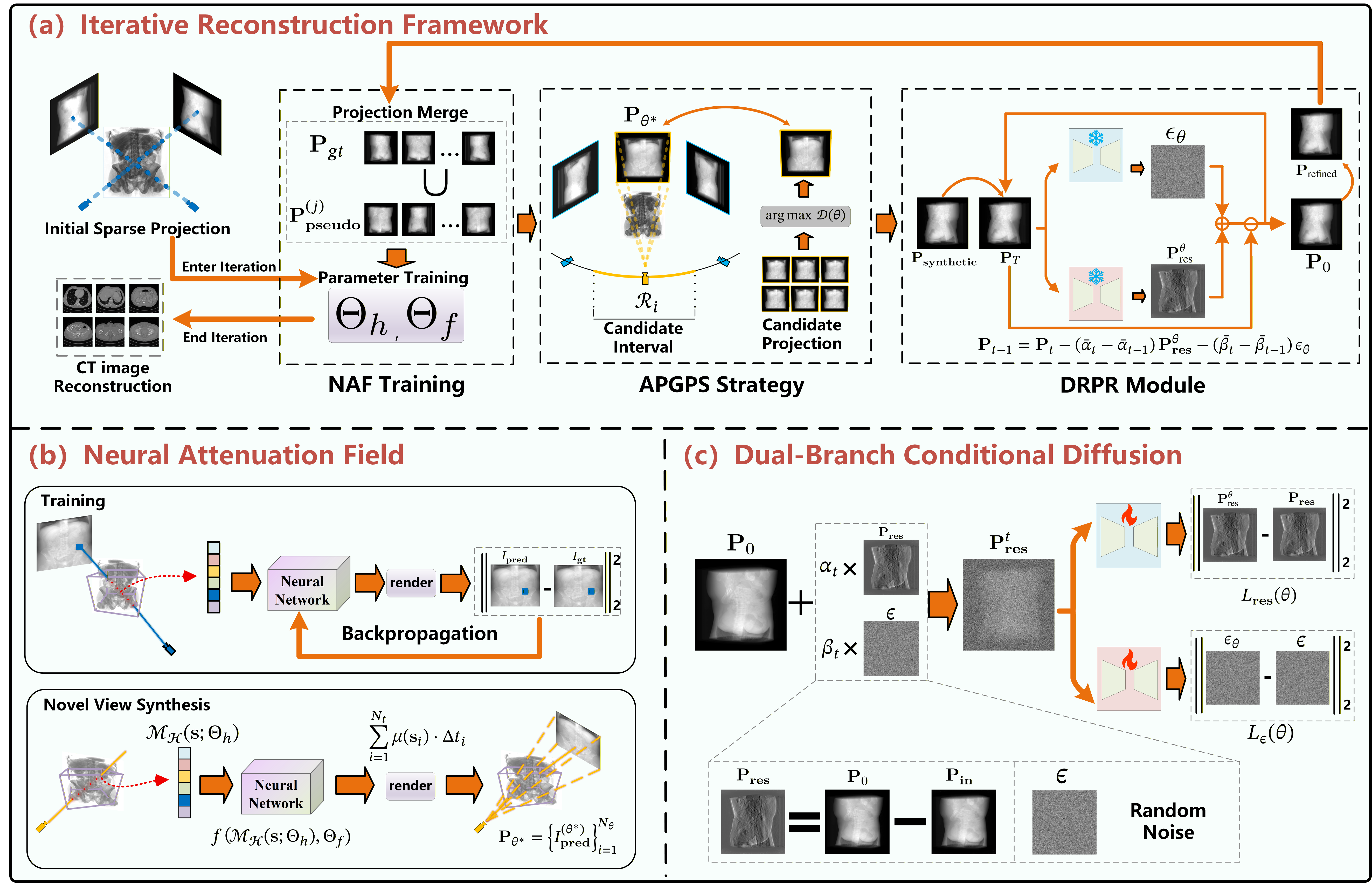}
    \caption{Overview of the proposed Diff-NAF framework. (a) Iterative reconstruction framework: starting from initial sparse-view projections, the NAF is trained and then enters an iterative process that includes novel-view synthesis, DR projection correction, pseudo-label merging, and NAF retraining, followed by CT reconstruction. (b) NAF: illustrates the training procedure and the synthesis of novel-view projections. (c) shows the training process of the dual-branch conditional diffusion model, which is the main component of the DRPR module.} 
    \label{fig3} 
\end{figure*}

This study proposes an iterative optimization framework that deeply integrates NAF with conditional diffusion models. It is designed for 3D CT reconstruction from ultra-sparse-view scans. The workflow of the framework is illustrated in Fig.~\ref{fig3}(a). The iterative process consists of three main parts: NAF training, new-view projection synthesis, and new projection refinement.

\textbf{NAF Training:} NAF is an implicit neural representation method that maps spatial coordinates to attenuation coefficients, enabling continuous parameterization of object structures. Its training process (Fig.~\ref{fig3}(b)) comprises three key steps: ray sampling, positional encoding, and attenuation coefficient prediction. Rays represent the paths of X-rays from source to detector. Each ray is parameterized as follows:
\begin{equation}
\mathbf{r}(t) = \mathbf{o} + t \cdot \mathbf{d}.
\label{eq:ray_equation}
\end{equation}

Here, $\mathbf{o}$ is the source position, $\mathbf{d}$ the unit direction vector, and $t$ the distance from the origin.

The ray sampling process selects points within the interval $t \in [t_0, t_1]$, where $t_0$ and $t_1$ mark ray entry and exit at the object boundary. The spatial coordinates of each sample point are
\begin{equation}
\mathbf{s}_i = \mathbf{o} + t_i \cdot \mathbf{d}, \quad i=1,\ldots,N_t,
\label{eq:sample_points}
\end{equation}
and the distance between adjacent samples is $\Delta t_i = t_{i+1} - t_i$.

The positional encoding is defined by
\begin{equation}
\mathcal{M}_{\mathcal{H}}(\mathbf{s};\Theta_h)
= \big[\mathcal{I}(\mathbf{H}_{1}),\ldots,\mathcal{I}(\mathbf{H}_{L})\big]^{T},
\label{eq:learned_hash_encoding}
\end{equation}
here $\mathcal{M}_{\mathcal{H}}$ concatenates $L$-level features for a spatial coordinate $\mathbf{s}$ using learnable hash tables $\mathbf{H}_l$ and an indexer $\mathcal{I}$; this adaptive design alleviates the spectral bias toward low frequencies~\cite{56,57}.

The attenuation prediction uses a deep network $f$ to map encoded spatial coordinates to attenuation coefficients:
\begin{equation}
\mu(\mathbf{s}) = f\!\Big( \mathcal{M}_{\mathcal{H}}(\mathbf{s}; \Theta_h), \Theta_f \Big).
\label{eq:overall_mapping}
\end{equation}

The total attenuation along a ray is approximated using sampled points $\mathbf{s}_i$ and spacings $\Delta t_i$ :
\begin{equation}
I_{\text{pred}} \approx \sum_{i=1}^{N_t} \mu(\mathbf{s}_i)\,\Delta t_i,
\label{eq:integral_approximation}
\end{equation}
where $N_t$ denotes the number of samples.

The resulting value matches the attenuation received by the detector and is used for supervision against the measured projections.

During training, the predicted intensity $I_{\text{pred}}$ is compared with the measured projection data $I_{\text{gt}}$. The loss function is
\begin{equation}
\mathcal{L} = \frac{1}{N_\theta}\sum_{i = 1}^{N_\theta}\big(I_{\text{gt}}^i - I_{\text{pred}}^i\big)^{2},
\label{eq:loss_function}
\end{equation}
where $I_{\text{gt}}^i$ is the measured projection for the $i$-th measurement, $I_{\text{pred}}^i$ is the corresponding prediction, and $N_\theta$ is the total number of projection measurements.

\textbf{New-view Projection Synthesis:} The trained NAF model predicts attenuation coefficients at any spatial point, enabling novel-view synthesis. An optimal projection view $\theta^{\ast} = \mathcal{A}(\theta)$ is first selected via the adaptive strategy $\mathcal{A}$, where $\theta$ denotes the available views and $\theta^\ast$ the target angular position. Based on this views, rays covering the entire detector area are generated. The ray generation for the novel view is:
\begin{equation}
\mathbf{r}^{(\theta^\ast)}(t) = \mathbf{o}_{\theta^\ast} + t \cdot \mathbf{d}_{\theta^\ast}.
\label{eq:novel_ray}
\end{equation}

Here, $\mathbf{o}_{\theta^\ast}$ is the source position for the novel views, and $\mathbf{d}_{\theta^\ast}$ is the direction vector determined by the projection geometry at view $\theta^\ast$.

Along each newly created ray, sampling is performed as
\begin{equation}
\mathbf{s}_i^{(\theta^\ast)} = \mathbf{o}_{\theta^\ast} + t_i \cdot \mathbf{d}_{\theta^\ast}, \quad i=1,\ldots,N_t,
\label{eq:novel_sampling}
\end{equation}
where $N_t$ is the number of sample points and $t_i \in [t_0,t_1]$ are the sampling parameters. The corresponding line integral for each novel ray is computed using the trained network as:
\begin{equation}
I_{\text{pred}}^{(\theta^\ast)} \approx \sum_{i=1}^{N_t} \mu\!\big(\mathbf{s}_i^{(\theta^\ast)}\big)\,\Delta t_i.
\label{eq:novel_integral}
\end{equation}

Finally, the complete projection data for the novel views are synthesized by accumulating line-integral values across the detector array, defined by
\begin{equation}
\mathbf{P}_{\theta^\ast} = \left\{ I_{\text{pred}}^{(\theta^\ast)} \right\}_{i=1}^{N_\theta^\ast},
\label{eq:complete_projection}
\end{equation}
where $N_\theta^\ast$ is the number of detector pixels and $I_{\text{pred}}^{(\theta^\ast)}$ denotes the predicted intensity for each X-ray path at view $\theta^\ast$.

\textbf{New Projection Refinement:} The synthesized novel views predicted by the model may contain subtle structural errors or artifacts. To improve projection quality, we use a DRPR module, defined by
\begin{equation}
\mathbf{P}_{\theta^\ast}^{\prime} = \mathcal{C}\!\big(\mathbf{P}_{\theta^\ast}\big).
\label{eq:correction_module}
\end{equation}

Here, $\mathcal{C}$ denotes the correction module, $\mathbf{P}_{\theta^\ast}$ is the initial synthetic projection for view $\theta^\ast$, and $\mathbf{P}_{\theta^\ast}^{\prime}$ is the corrected projection output.

To explicitly distinguish between original and augmented data, we define the initial sparse projection set as
\begin{equation}
\mathbf{P}_{\text{gt}} = \big\{ \mathbf{P}_{y}^{(i)} \big\}_{i=1}^{N_0},
\label{eq:data_gt_init_reparation}
\end{equation}
and let $\mathbf{P}_{\text{pseudo}}^{(k)}$ denote the set of corrected projections added at iteration $k$. Accordingly, the training set at iteration $k+1$ is updated as
\begin{equation}
\mathbf{P}^{(k+1)} = \mathbf{P}^{(k)} \cup \mathbf{P}_{\text{pseudo}}^{(k+1)}.
\label{eq:data_augmentation}
\end{equation}

Here, $\mathbf{P}^{(0)} = \mathbf{P}_{\text{gt}}$, $I_{\text{gt}} \in \mathbf{P}_{\text{gt}}$ denotes a real measured projection, whereas $I_{\text{pseudo}} \in \mathbf{P}_{\text{pseudo}}^{(j)}$ represents a corrected projection used as a pseudo-label.

\textbf{Iteration:} By iterating this process, we progressively expand the training dataset from an initially ultra-sparse state to a gradually sufficient one. At each iteration, the optimization objective is defined over both the original projection set $\mathbf{P}_{\text{gt}}$ and the accumulated corrected sets $\mathbf{P}_{\text{pseudo}}^{(k)}$. To balance their contributions, we assign a weight $w_1$ to real measured projections and a weight $w_2$ to corrected projections; the objective is
\begin{equation}
\begin{aligned}
\min_{\Theta_f,\Theta_h}\;
&\sum_{I_{\text{gt}} \in \mathbf{P}_{\text{gt}}}
w_{1}\,\big( I_{\text{gt}} - I_{\text{pred}}(\Theta_f,\Theta_h) \big)^2 \\
&\qquad + \sum_{I_{\text{pseudo}} \in \mathbf{P}_{\text{pseudo}}^{(k)}}
w_{2}\,\big( I_{\text{pseudo}} - I_{\text{pred}}(\Theta_f,\Theta_h) \big)^2 .
\end{aligned}
\label{eq:iterative_objective}
\end{equation}

Here, $\Theta_f$ and $\Theta_h$ denote the neural field and hash encoding parameters of the NAF model, respectively.

\begin{algorithm}[!htbp]
\footnotesize
\caption{Diff-NAF Iterative Optimization Framework}
\label{alg:naf_optimization}
\begin{algorithmic}[1]

\Require Original projection set $\mathbf{P}_{\text{gt}} = \{(\theta_i, \mathbf{P}_{\theta_i})\}_{i=1}^{N_0}$, correction module $\mathcal{C}$, initial parameters $(\Theta_f^{(0)}, \Theta_h^{(0)})$, number of iterations $K$, weights $(w_1,w_2)$
\Ensure Optimized neural field $f(\cdot;\Theta_f^\star)$ and hash encoding $\mathcal{M}_{\mathcal{H}}(\cdot;\Theta_h^\star)$

\State \textbf{Initialize / NAF Training:}
Train $f(\cdot;\Theta_f^{(0)})$ with $\mathcal{M}_{\mathcal{H}}(\cdot;\Theta_h^{(0)})$ on $\mathbf{P}_{\text{gt}}$ using the loss
$\sum_{I_{\text{gt}}\in \mathbf{P}_{\text{gt}}}
\big( I_{\text{gt}} - I_{\text{pred}}(\Theta_f^{(0)},\Theta_h^{(0)}) \big)^2$.

\For{$k = 1$ \textbf{to} $K$}
    \For{each selected target view $\theta^\ast$}
        \State Generate per-pixel rays $\mathbf{r}^{(\theta^\ast)}(t) = \mathbf{o}_{\theta^\ast} + t \cdot \mathbf{d}_{\theta^\ast}$.
        \State Sample along each ray: $\mathbf{s}_i^{(\theta^\ast)} = \mathbf{o}_{\theta^\ast} + t_i \cdot \mathbf{d}_{\theta^\ast}$.
        \State Predict attenuation:
        \Statex \quad $\mu(\mathbf{s}_i^{(\theta^\ast)}) \gets f\!\big(\mathcal{M}_{\mathcal{H}}(\mathbf{s}_i^{(\theta^\ast)};\Theta_h^{(k-1)}),\,\Theta_f^{(k-1)}\big)$.
        \State Ray integral: $I_{\text{pred}}^{(\theta^\ast)} \gets \sum_{i=1}^{N_t} \mu(\mathbf{s}_i^{(\theta^\ast)})\,\Delta t_i$.
        \State Form projection: $\mathbf{P}_{\theta^\ast} \gets \{\, I_{\text{pred}}^{(\theta^\ast)} \,\}_{u=1}^{N_\theta}$.
    \EndFor

    \State Obtain corrected projection for each $\theta^\ast$: $\mathbf{P}_{\theta^\ast}^{\prime} \gets \mathcal{C}(\mathbf{P}_{\theta^\ast})$.
    \State Define the newly added pseudo set at iteration $k$:
    \Statex \quad $\mathbf{P}_{\text{pseudo}}^{(k)} \gets \{(\theta^\ast, \mathbf{P}_{\theta^\ast}^{\prime})\}$.
    \State \textbf{Projection-Set Augmentation:} Update
    \Statex \quad $\mathbf{P}^{(k)} \gets \mathbf{P}^{(k-1)} \cup \mathbf{P}_{\text{pseudo}}^{(k)}$.
    \State Update parameters by minimizing the weighted loss:
    {\small
    \[
      \begin{aligned}
      (\Theta_f^{(k)}, \Theta_h^{(k)}) \gets &\arg\min_{\Theta_f,\Theta_h}\;
      \sum_{I_{\text{gt}} \in \mathbf{P}_{\text{gt}}}
      w_{1}\,\big( I_{\text{gt}} - I_{\text{pred}}(\Theta_f,\Theta_h) \big)^2 \\
      &\qquad + \sum_{I_{\text{pseudo}} \in \mathbf{P}_{\text{pseudo}}^{(k)}}
      w_{2}\,\big( I_{\text{pseudo}} - I_{\text{pred}}(\Theta_f,\Theta_h) \big)^2 .
      \end{aligned}
    \]
    }
\EndFor

\State \Return $f(\cdot;\Theta_f^{(K)})$ and $\mathcal{M}_{\mathcal{H}}(\cdot;\Theta_h^{(K)})$.
\end{algorithmic}
\end{algorithm}

\subsection{Diffusion-driven Reuse Projection Refinement Module}
In the iterative optimization pipeline, projection refinement is a critical step. DRPR module employs a dual-branch diffusion architecture to address the lack of physical constraints in DR projection refinement. Here, $\mathbf{P}_{\text{in}}$ denotes conditional projection, $\epsilon$ is random noise, and the degraded projection is defined as $\mathbf{P}_T = \mathbf{P}_{\text{in}} + \epsilon$. This forward diffusion simultaneously models structural optimization and progressive noise injection.

A single forward process step of the diffusion model is written as
\begin{align}
\mathbf{P}_t = \mathbf{P}_{t-1} + \mathbf{P}_{\text{res}}^t .
\label{eq:single_forward_process}
\end{align}

Here, $\mathbf{P}_{\text{res}}^t$ represents the directional mean shift from state $\mathbf{P}_{t-1}$ to state $\mathbf{P}_t$ with random perturbation.

The residual distribution is specified as
\begin{align}
\mathbf{P}_{\text{res}}^t \sim \mathcal{N}\!\bigl(\alpha_t \mathbf{P}_{\text{res}},\, \beta_t \epsilon\bigr) ,
\label{eq:residual_distribution}
\end{align}
where the residual $\mathbf{P}_{\text{res}}$ in $\mathbf{P}_{\text{res}}^t$ is the difference between $\mathbf{P}_{\text{in}}$ and $\mathbf{P}_0$, and two independent coefficient schedules $\alpha_t$ and $\beta_t$ control the residual and noise diffusion, respectively.

The transformation process from $\mathbf{P}_0$ to $\mathbf{P}_t$ is stated as
\begin{align}
\mathbf{P}_t &= \mathbf{P}_{t-1} + \alpha_t \mathbf{P}_{\text{res}} + \beta_t \boldsymbol{\epsilon}_{t-1} \nonumber \\
             &= \mathbf{P}_0 + \bar{\alpha}_t \mathbf{P}_{\text{res}} + \bar{\beta}_t \boldsymbol{\epsilon} ,
\label{eq:detailed_process}
\end{align}
in which $\mathcal{N}$ denotes the Gaussian distribution, $\alpha_t \mathbf{P}_{\text{res}}$ represents the deterministic trend of the residual with $\alpha_t$ being the weight coefficient and $\mathbf{P}_{\text{res}}$ being the target residual, $\beta_t$ controls the noise scale, and $\epsilon$ is the random noise.

Based on the above derivation, the joint probability distribution in the forward process factorizes into conditional Gaussians as
\begin{equation}
\begin{aligned}
q\!\left(\mathbf{P}_{1:T} \mid \mathbf{P}_0, \mathbf{P}_{\text{res}}\right)
&= \prod_{t=1}^T q\!\left(\mathbf{P}_t \mid \mathbf{P}_{t-1}, \mathbf{P}_{\text{res}}\right), \\
q\!\left(\mathbf{P}_t \mid \mathbf{P}_{t-1}, \mathbf{P}_{\text{res}}\right)
&= \mathcal{N}\!\left(\mathbf{P}_t;\, \mathbf{P}_{t-1} + \alpha_t \mathbf{P}_{\text{res}},\, \beta_t \epsilon \right) .
\end{aligned}
\label{eq:joint_distribution}
\end{equation}

Here, $\alpha_t$ and $\beta_t$ control the residual and noise scales at each step.

The reverse process from $\mathbf{P}_T$ to $\mathbf{P}_0$ requires estimation of both residual and noise. A residual prediction model $\mathbf{P}_{\text{res}}^{\theta}(\mathbf{P}_t, t, \mathbf{P}_{\text{in}})$ and a noise prediction model $\epsilon_{\theta}(\mathbf{P}_t, t, \mathbf{P}_{\text{in}})$ are trained, and the single-step sampling is expressed as
\begin{equation}
\mathbf{P}_{t-1} = \mathbf{P}_t
- \bigl(\bar{\alpha}_t - \bar{\alpha}_{t-1}\bigr)\, \mathbf{P}_{\text{res}}^{\theta}
- \bigl(\bar{\beta}_t - \bar{\beta}_{t-1}\bigr)\, \epsilon_{\theta} .
\label{eq:reverse_single_step}
\end{equation}

The training objective functions for the residual prediction model and the noise prediction model are written as
\begin{equation}
\begin{aligned}
L_{\text{res}}(\theta) &= \lambda_{\text{res}}\left\|\mathbf{P}_{\text{res}}
- \mathbf{P}_{\text{res}}^{\theta}\!\left(\mathbf{P}_{t}, t, \mathbf{P}_{\text{in}}\right)\right\|^{2}, \\
L_{\epsilon}(\theta) &= \lambda_{\epsilon}\left\|\epsilon
- \epsilon_{\theta}\!\left(\mathbf{P}_{t}, t, \mathbf{P}_{\text{in}}\right)\right\|^{2}.
\end{aligned}
\label{eq:loss_res_noise}
\end{equation}

Another challenge encountered in the iterative framework is the inconsistency in data flow, which stems from the mismatch between the projection data generated by the NAF and the input–output characteristics required by the diffusion model. A DRAT method is designed to resolve this conflict by achieving dynamic data range adaptation through invertible linear transformations; the forward affine mapping is
\begin{align}
\mathbf{P}_{\text{T}} = \boldsymbol{\Gamma}\, \mathbf{P}_{\text{synthetic}} + \mathbf{b} ,
\label{eq:positive_linear}
\end{align}
where $\boldsymbol{\Gamma}=\operatorname{diag}(\gamma_1,\ldots,\gamma_m)$ is a diagonal scaling matrix and $\mathbf{b}=[b_1,\ldots,b_m]^{\top}$ is the shift vector. A positive regularization term ensures numerical stability by preventing singular scaling. For the refined projection $\mathbf{P}_0$, the exact inverse affine mapping recovers the raw scale:
\begin{align}
\mathbf{P}_{\text{refined}} = \boldsymbol{\Gamma}^{-1}\bigl(\mathbf{P}_0 - \mathbf{b}\bigr) .
\label{eq:reverse_linear}
\end{align}

Before refinement, the dynamic-range parameters of $\mathbf{P}_{\text{synthetic}}$ are estimated from the dataset, and then the affine linear mapping above is applied with the additional regularization term to stabilize the scaling coefficients and avoid degenerate transformations.

\begin{figure}[htbp]
    \centering
    \includegraphics[width=\columnwidth]{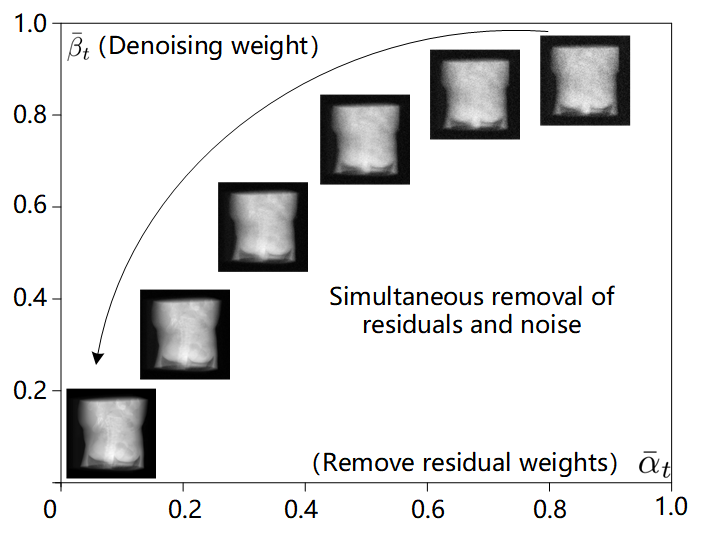}
    \caption{Illustration of the dual-branch diffusion correction process. 
The vertical axis ($\bar{\beta}_t$) denotes the denoising weight, and the horizontal axis ($\bar{\alpha}_t$) denotes the residual removal weight. 
The curve shows the evolution of DR correction, demonstrating the simultaneous removal of residuals and noise during the diffusion process.}    
    \label{fig4} 
\end{figure}

\subsection{Angle-Prior Guided Projection Synthesis}
In the iterative framework, the selection strategy for novel projection views plays a significant role in the final reconstruction quality. This study proposes an adaptive projection selection strategy. Let the known set of projection views be $\Theta = \{\theta_1, \theta_2, \ldots, \theta_N\}$ with $\theta_i < \theta_{i+1}$. Define the candidate interval $\mathcal{R}_i$ by
\begin{equation}
\mathcal{R}_i = \Biggl[\frac{\theta_i+\theta_{i+1}}{2} - \frac{\Delta\theta}{a},\;
\frac{\theta_i+\theta_{i+1}}{2} + \frac{\Delta\theta}{a}\Biggr],
\label{eq:angle_strategy}
\end{equation}
where $\Delta\theta = \theta_{i+1} - \theta_i$. This interval avoids the “high-confidence regions” close to known views. Based on past experience, setting $a=4$ is a more effective choice.

For each candidate view $\theta \in \mathcal{R}_i$, compute the gradient-based dissimilarity with respect to its neighboring projections as
\begin{equation}
\resizebox{\columnwidth}{!}{$
\begin{aligned}
\mathcal{D}(\theta) 
&= \frac{1}{HW}\sum_{y=1}^{H} \sum_{x=1}^{W}
\left\|
\nabla \mathbf{P}_\theta(x,y) \;-\;
\frac{\nabla \mathbf{P}_{\theta_i}(x,y)+\nabla \mathbf{P}_{\theta_{i+1}}(x,y)}{2}
\right\|_2 .
\end{aligned}
$}
\label{eq:similarity_calculation}
\end{equation}

Finally, select the novel projection views as $\theta^\ast = \arg\max_{\theta \in \mathcal{R}_i} \mathcal{D}(\theta)$, making the strategy fully automatic and improving robustness while maintaining efficiency, thereby enhancing the effectiveness of the joint training framework.

\section{EXPERIMENTS}\label{sec:experiments}
\subsection{Datasets and Evaluation Metrics}
\begin{table*}[t]
\centering
\caption{Quantitative CT reconstruction results under the 50-view setting. Entries are PSNR/SSIM on six datasets.}
\label{tab:50_CT}
\begin{tabular*}{\textwidth}{@{\extracolsep{\fill}}c c c c c c c@{}}
  \toprule
  Method/Dataset & L067 & L096 & Box & Foot & Head & Jaw \\
  \midrule
  FDK~\cite{FBP}        & 22.15/0.7015 & 22.91/0.7208 & 26.10/0.7841 & 24.35/0.7133 & 26.97/0.8107 & 24.19/0.6821 \\
  SART~\cite{SART}       & 31.42/0.9058 & 31.39/0.9097 & 33.11/0.9471 & 30.22/0.9163 & 34.75/0.9535 & 30.34/0.8944 \\
  InTomo~\cite{32}     & 32.53/0.9387 & 33.42/0.9434 & 33.65/0.9564 & 30.86/0.9218 & 35.73/0.9688 & 32.55/0.9131 \\
  NeRF~\cite{30}       & 32.79/0.9422 & 34.78/0.9579 & 35.40/0.9723 & 31.18/0.9269 & 36.81/0.9777 & 34.12/0.9368 \\
  NeAT~\cite{51}       & 32.74/0.9421 & 33.84/0.9493 & 34.86/0.9684 & 31.19/0.9272 & 35.98/0.9713 & 33.70/0.9308 \\
  NAF~\cite{34}        & 34.91/0.9576 & 35.41/0.9617 & 36.59/0.9764 & 31.70/0.9331 & 37.83/0.9817 & 34.40/0.9387 \\
  SAX-NeRF~\cite{54}   & 35.15/0.9626 & 37.79/0.9770 & 38.61/0.9864 & 31.59/0.9332 & 39.40/0.9881 & 34.76/0.9473 \\
  Diff-NAF             & \textbf{37.21/0.9780} & \textbf{39.13/0.9833} & \textbf{39.76/0.9898} & \textbf{32.09/0.9414} & \textbf{41.26/0.9918} & \textbf{36.49/0.9647} \\
  \bottomrule
\end{tabular*}
\end{table*}

\begin{figure*}[t]
    \centering
    \includegraphics[scale=0.43]{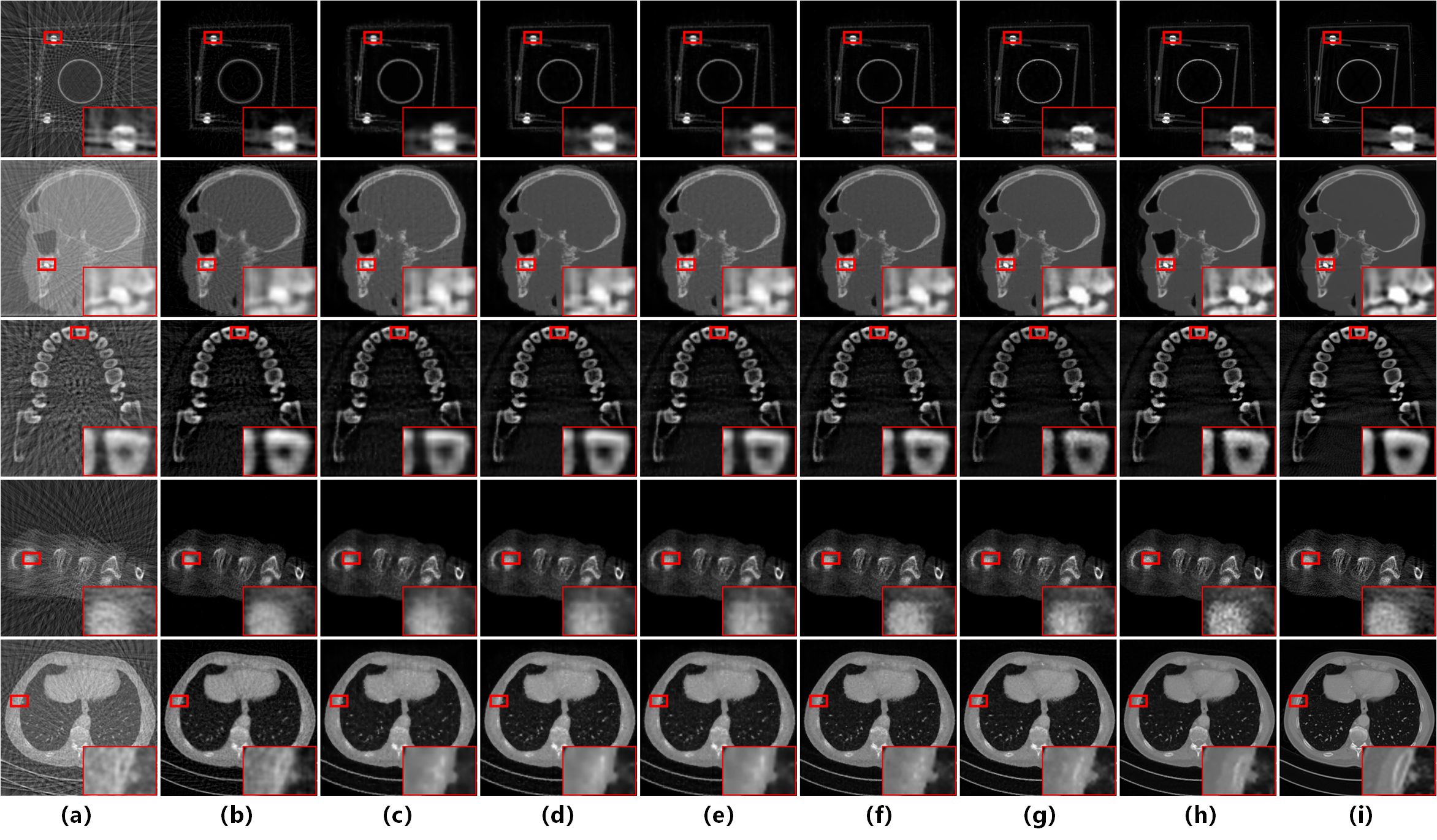}
    \caption{Qualitative CT reconstruction comparison under the 50-view setting. Representative slices reconstructed by eight methods are shown. The vertical axis lists five datasets: Box, Head, Jaw, Foot, and AAPM. The horizontal axis, from (a) to (i), corresponds to FDK, SART, InTomo, NeRF, NeAT, NAF, SAX-NeRF, Diff-NAF, and the ground truths.}
    \label{fig:50_CT} 
\end{figure*}

This study uses five public datasets and one real projection dataset. The public datasets include simulated chest CT volumes from the Mayo Clinic AAPM Low-Dose CT Grand Challenge \cite{63} and four volumes (jaw, head, box, foot) from the Open Science Visualization Dataset \cite{64}. The real dataset consists of in-house DR projections of a chicken wing bone placed in a test tube. For the simulated data, each 3D volume is $256\times256\times256$ with $1.0\,\mathrm{mm}$ isotropic voxels. DR projections are generated using a flat-panel detector of $512\times512$ pixels with $1.0\,\mathrm{mm}$ pixel pitch, and the source–object and source–detector distances (SOD/SDD) are set to $1000/1500\,\mathrm{mm}$. For the chicken-wing–bone data, the detector resolution is $1536\times1536$ with $0.085\,\mathrm{mm}\times0.085\,\mathrm{mm}$ pixel size, SOD/SDD are $300/325\,\mathrm{mm}$, and reconstructed volumes are $768\times768\times768$ with $0.098\,\mathrm{mm}$ isotropic voxels. For the chicken-wing dataset, five specimens are available: four specimens are used to train the DRPR diffusion model, and one held-out specimen is reserved exclusively for evaluating the final framework; the training and inference view sets remain non-overlapping.

The framework trains a diffusion model inside the DRPR module using DR projections. For all datasets, 720-view DR projections are used for training, and the diffusion model is trained for 1000 epochs. For the AAPM data, 10 volumes are available; 8 volumes generate DR projections for training the DRPR diffusion model, and the remaining 2 volumes are used for iterative reconstruction. For the jaw, head, box, and foot datasets, each provides a single volume. Due to limited data, the same 3D volume is used for training the diffusion model and for iterative reconstruction; however, the training and inference view sets are non-overlapping, so the DR projections used for training and those refined during iterative reconstruction come from different views. 

Iterative reconstruction is evaluated under two settings: ultra-sparse 20-view and sparse 50-view. The density prediction network is optimized for 1,500 gradient steps in each outer iteration. All experiments are run on RTX 5070 Ti and A6000 GPUs. This study employs two quantitative metrics, SSIM and PSNR. For the imaging results, the metrics are computed between the reconstructed 3D volume and the ground-truth 3D volume. For more in-depth study and research, the source code is at: \href{https://github.com/yqx7150/Diff-NAF}{https://github.com/yqx7150/Diff-NAF}.

\subsection{Comparative Experiments}
\begin{table*}[t]
  \centering
  \caption{Quantitative CT reconstruction results under the 20-view setting. Entries are PSNR/SSIM on six datasets.}
  \label{tab:20_CT}
  \begin{tabular*}{\textwidth}{@{\extracolsep{\fill}}c c c c c c c@{}}
    \toprule
    Method/Dataset & L067 & L096 & Box & Foot & Head & Jaw \\
    \midrule
    FDK        & 21.64/0.6012 & 20.04/0.6100 & 23.41/0.5283 & 20.17/0.5994 & 21.02/0.5816 & 21.36/0.5829 \\
    SART       & 26.54/0.7554 & 28.47/0.7256 & 29.85/0.8620 & 28.41/0.8321 & 28.19/0.8350 & 27.91/0.8097 \\
    InTomo     & 28.74/0.8726 & 29.07/0.8105 & 29.03/0.8342 & 29.40/0.9042 & 29.27/0.8355 & 28.79/0.8189 \\
    NeRF       & 26.87/0.8282 & 28.02/0.8374 & 30.41/0.9003 & 28.52/0.8928 & 29.30/0.8857 & 27.87/0.7908 \\
    NeAT       & 27.59/0.8416 & 29.68/0.8808 & 31.88/0.9313 & 29.38/0.9018 & 31.45/0.9210 & 29.54/0.8383 \\
    NAF        & 29.18/0.8813 & 31.10/0.9075 & 32.93/0.9455 & 29.64/0.9069 & 31.30/0.9215 & 30.06/0.8517 \\
    SAX-NeRF   & 28.45/0.8587 & 32.27/0.9267 & 33.26/0.9522 & 29.45/0.9070 & 32.95/0.9515 & 29.97/0.8514 \\
    Diff-NAF   & \textbf{32.17/0.9308} & \textbf{34.87/0.9561} & \textbf{35.23/0.9708} & \textbf{30.50/0.9196} & \textbf{34.90/0.9651} & \textbf{31.99/0.9011} \\
    \bottomrule
  \end{tabular*}
\end{table*}
\begin{figure*}[t]
    \centering
    \includegraphics[scale=0.43]{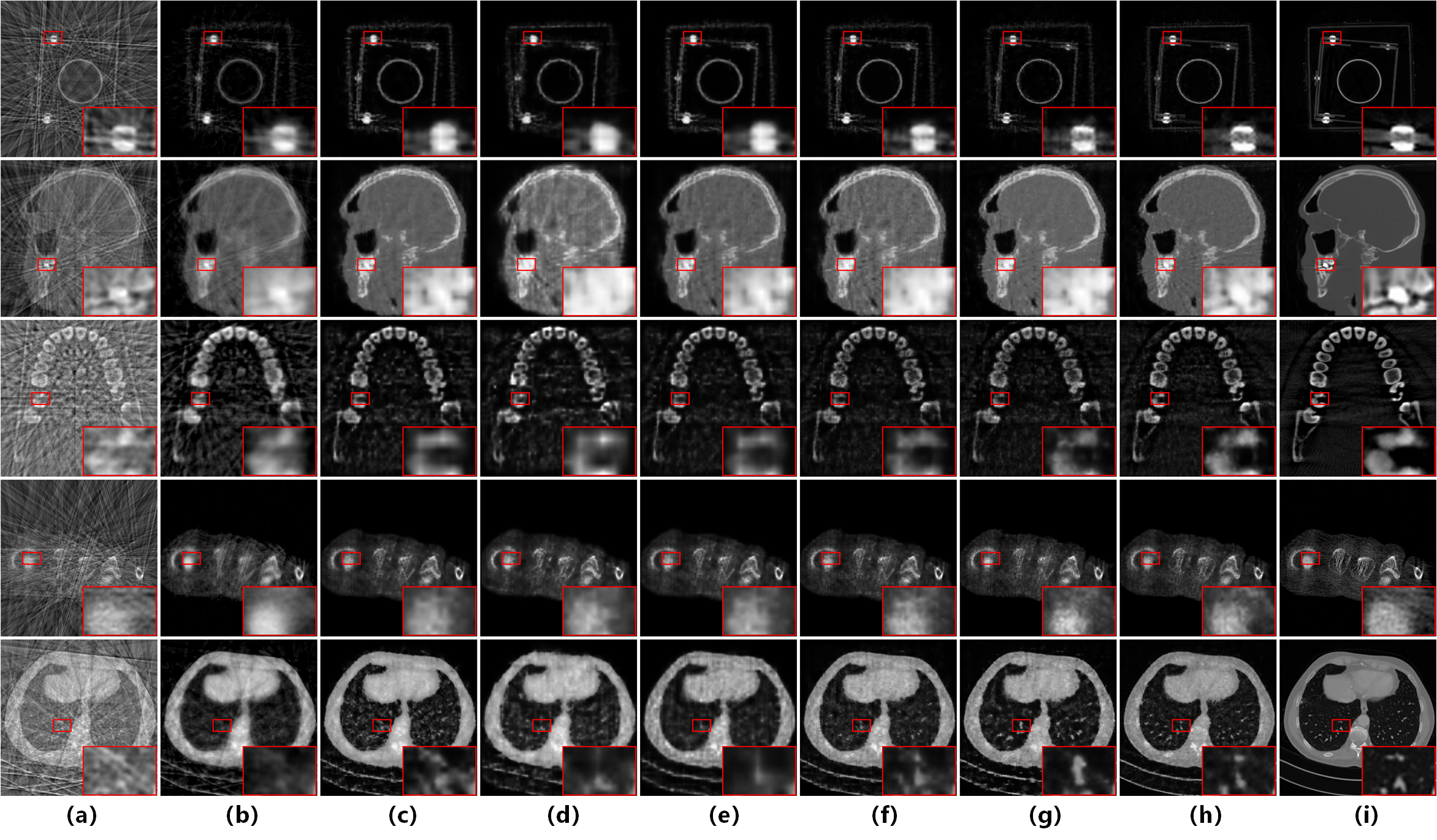}
    \caption{Qualitative CT reconstruction comparison under the 20-view setting. Representative slices reconstructed by eight methods are shown. The vertical axis lists five datasets: Box, Head, Jaw, Foot, and AAPM. The horizontal axis, from (a) to (i), corresponds to FDK, SART, InTomo, NeRF, NeAT, NAF, SAX-NeRF, Diff-NAF, and the ground truths.} 
    \label{fig:20_CT} 
\end{figure*}

For 3D CT reconstruction in the 50-view setting, we evaluate against the classical baselines FDK and SART, as well as InTomo, NeRF, NeAT, NAF, and SAX-NeRF, alongside the methods described above. PSNR/SSIM results are summarized in Table~\ref{tab:50_CT}. The proposed method outperforms all competitors on every dataset, achieving gains of 2.06\,dB and 1.23\,dB on AAPM L067 and L096, and 0.50\,dB, 1.86\,dB, 1.73\,dB, and 1.15\,dB on foot, head, jaw, and box, respectively. As illustrated in Fig.~\ref{fig:50_CT}, the reconstructions exhibit stronger noise suppression and higher structural fidelity.

For ultra-sparse (20-view) 3D CT reconstruction, Table~\ref{tab:20_CT} reports the PSNR/SSIM results. The proposed method achieves the best performance across all datasets, with gains of 2.99\,dB and 2.60\,dB on AAPM L067 and L096, and 1.05\,dB, 1.95\,dB, 2.02\,dB, and 1.97\,dB on foot, head, jaw, and box, respectively. The visualizations in Fig.~\ref{fig:20_CT} further demonstrate improved noise suppression and enhanced recovery of structural details under ultra-sparse sampling.

\begin{figure}[htbp]
    \centering
    \includegraphics[width=\columnwidth]{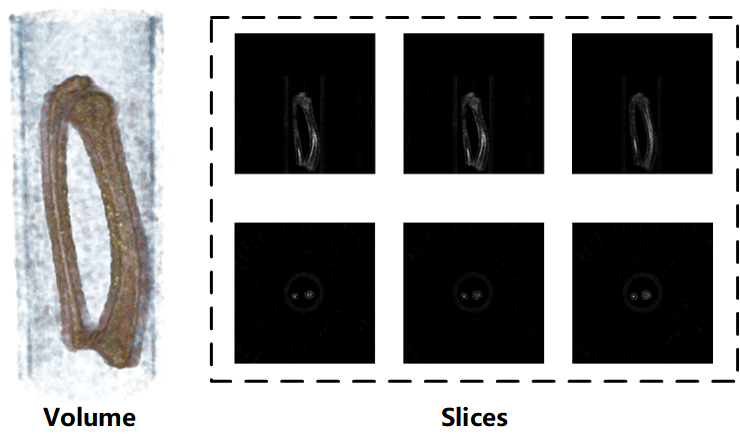}
    \caption{CT reconstructions of a chicken wing placed in a glass tube under the 20-views setting. The left panel shows a 3D volume rendering, and the right panel shows several representative axial and coronal slices reconstructed using the proposed Diff-NAF method.}
    \label{fig:jichi} 
\end{figure}

Beyond the public datasets described above, we also acquired real projections with our own hardware for reconstruction. Fig.~\ref{fig:jichi} shows a real dataset obtained by placing a chicken wing inside a glass test tube. Using a multi-source stationary CT system, we collected 20 projection views uniformly over 360°, which yielded measured projections from the physical object. The reconstruction produced by our method is presented in Fig.~\ref{fig:jichi}. Even with ultra-sparse-view sampling with only 20 views, streak artifacts and other sparse-view artifacts are greatly reduced, and the anatomy is faithfully recovered. 

\subsection{Ablation Study}
\begin{figure}[htbp]
    \centering
    \includegraphics[width=\columnwidth]{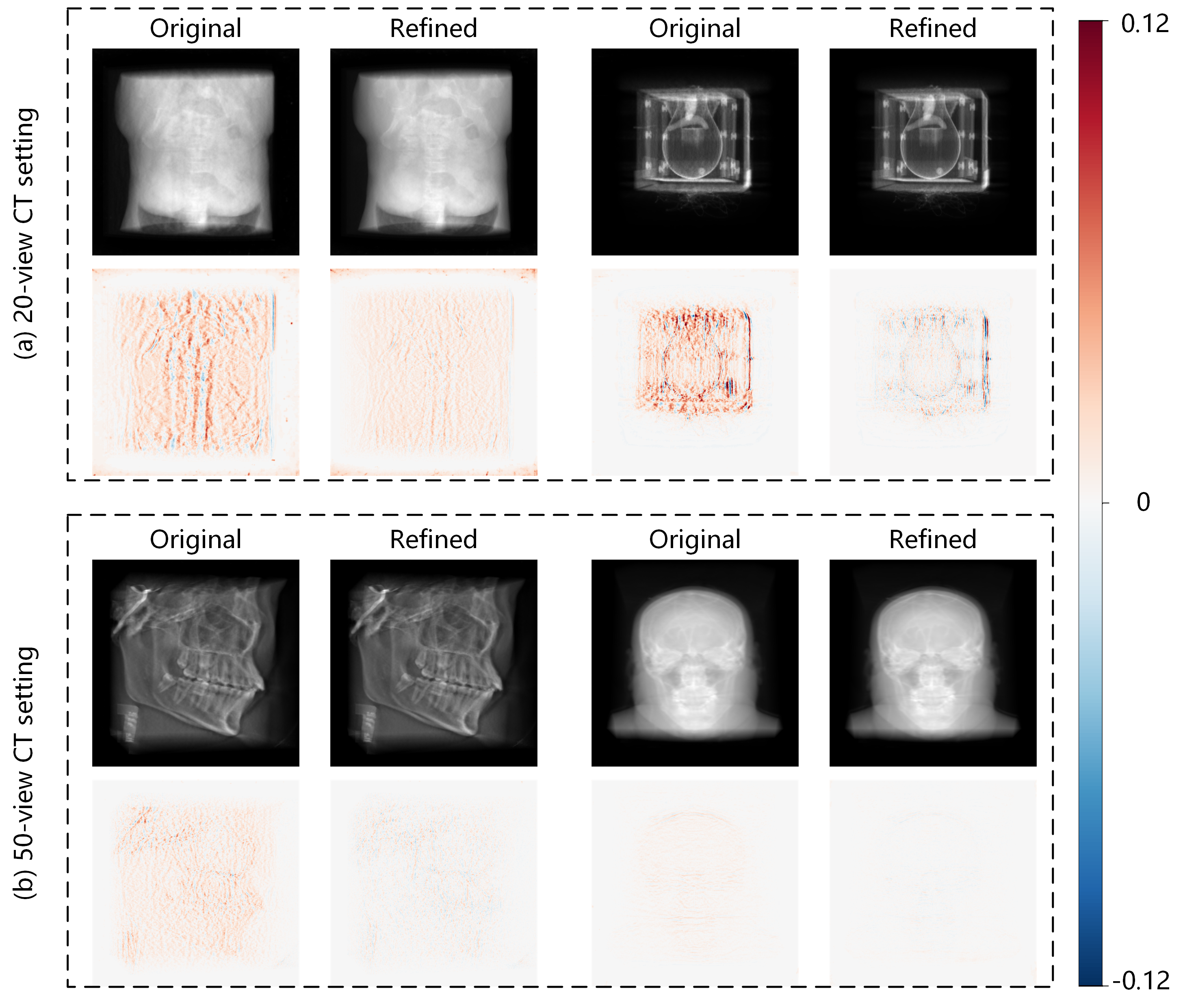}
    \caption{Residual analysis of novel-view DR projection synthesis under the 20-view setting. Original denotes DR projections directly generated by SAX-NeRF, and refined denotes DR projections refined by the DRPR module. In addition, (a) shows results for AAPM and box under the 20-view condition, (b) shows results for jaw and head under the 50-view condition. The right side displays the residual values, computed as the difference between the normalized DR projections and the ground-truth DR projections.}
    \label{fig:DR_res}
\end{figure}

\textbf{DRPR Model:} Within the iterative framework, the DRPR model refines the newly generated DR projections. Fig.~\ref{fig:DR_res} presents the residuals with respect to the ground-truth projections for the DR results before and after refinement by the DRPR module. Panel~(a) corresponds to the 20-view experiment. We select the optimal iteration step and, at that step, obtain the DR projections before and after refinement; after normalization, these are compared with the ground-truth projections to compute the residuals. Likewise, panel~(b) shows the results under the 50-view setting. As illustrated, under both experimental conditions the DRPR model effectively refines the DR projections and reduces the errors.

\begin{figure}[t]
    \centering
    \includegraphics[width=\columnwidth]{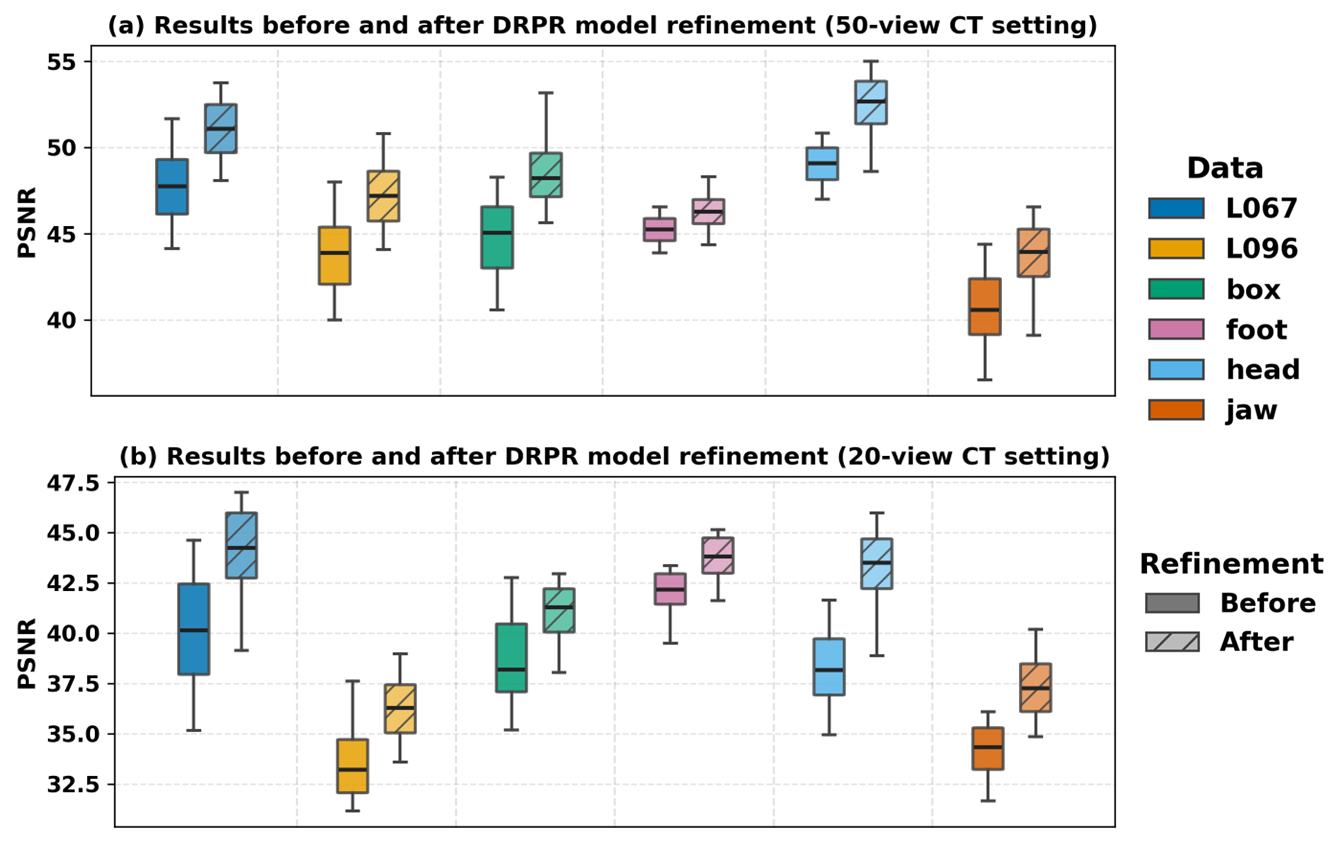}
    \caption{Box plots of novel-view DR projection refinement. Statistics are computed on the DR projection samples generated and optimized at the optimal iteration step for each dataset. Panel (a) reports the PSNR of DR projections before and after refinement across six datasets under the 50-view setting; likewise, panel (b) reports the results under the 20-view setting.}
    \label{fig:box plots}
\end{figure}

Fig.~\ref{fig:box plots} shows the box plots of these projections before and after refinement. Panels~(a) and~(b) correspond to the 50-view and 20-view experimental settings, respectively. It can be seen that, across both settings and for every dataset, the DRPR model effectively refines the DR projections.

To verify the generality of the proposed iterative self-supervised framework, we conducted ablation experiments using the AAPM L067 and L096 3D samples. Reported results are the average performance across the two samples. The evaluation compares three configurations based on three representative INRs: NeRF, NAF, and SAX-NeRF. Specifically, (1) Baseline denotes the original model without our framework; (2) w/o DRPR applies the iterative optimization but excludes DRPR module; and (3) Full Framework represents the complete iterative optimization framework proposed in this study. Quantitative results are summarized in Table~\ref{tab:ablation}.

\begin{table}[htbp]
  \centering
  \caption{Ablation study on the AAPM dataset (averaged over L067 and L096). Comparison of baseline, iteration without DRPR, and the full iterative framework. Metrics are reported as PSNR/SSIM.}
  \label{tab:ablation}
  \begin{tabular}{c c c c}
    \toprule
    Method & Baseline & w/o DRPR & Full Framework \\
    \midrule
    NeRF      & 33.79/0.9500 & 33.84/0.9510 & \textbf{35.24/0.9694} \\
    NAF       & 35.16/0.9597 & 35.25/0.9594 & \textbf{36.17/0.9778} \\
    SAX-NeRF  & 36.47/0.9698 & 36.50/0.9703 & \textbf{38.17/0.9807} \\
    \bottomrule
  \end{tabular}
\end{table}

\begin{figure}[htbp]
\centering
\includegraphics[width=\columnwidth]{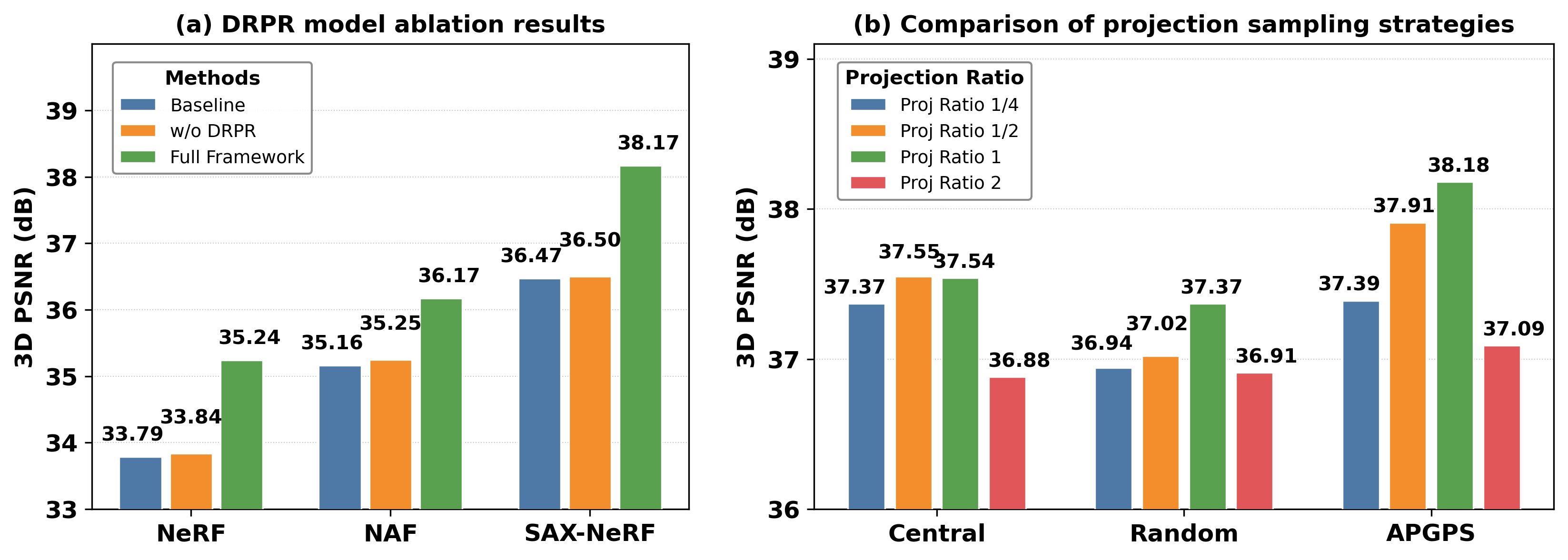}
\caption{
Ablation study results. 
Left: performance of the DRPR module across different baseline models, comparing Baseline, w/o DRPR, and Full Framework. 
Right: quantitative analysis of different sampling strategies (Central, Random, and APGPS) under four projection ratios.
}
\label{fig:ablation_bar}
\end{figure}

The ablation results of the DRPR module across different baseline models are visualized left figure in Fig.~\ref{fig:ablation_bar}. The figure presents comparative performance in terms of PSNR/SSIM under three configurations: the baseline model, the iterative optimization without DRPR, and the full iterative framework including DRPR. Without incorporating the iterative framework, each baseline model quickly reaches its performance limit after standard training. Introducing the iterative optimization alone (w/o DRPR) yields marginal improvement, indicating that iterative fine-tuning without projection refinement contributes little to performance enhancement. In contrast, when the DRPR module is integrated into the iterative process (Full Framework), all baseline models exhibit significant and consistent performance gains. This result highlights that the proposed DRPR-driven iterative mechanism effectively breaks the convergence bottleneck of existing models and provides stable improvement across different architectures.

\textbf{Sampling Strategy:} This study further performs an ablation experiment on the projection selection strategy within the proposed iterative training framework. In our iterative process, the NAF network is trained to perform novel-view synthesis, the generated DR projections are refined through the DRPR module, and the refined projections are then reintroduced into NAF training for iterative optimization. The projection selection strategy determines how new views are selected for synthesis in each iteration, which directly affects the diversity and quality of the training data. We compare three selection strategies: Central, Random, and our APGPS. The Central strategy selects new views at the midpoint between two known angles, while the Random strategy randomly samples views. In contrast, APGPS adaptively selects projection angles based on the geometric relationships among existing views, similar to central sampling but with learned offset adjustments to prioritize underrepresented viewing directions.

As shown in Table~\ref{tab:apgps}, each strategy is evaluated under four projection quantity ratios: 1/4, 1/2, 1, and 2, representing the number of synthesized novel views relative to the original 50-view training set. For instance, a ratio of 1/2 corresponds to generating 25 new projections, while 2 corresponds to doubling the number of projections by inserting two new angles between each known pair. Results are averaged over the AAPM L067 and L096 3D samples. The results show that APGPS consistently outperforms the other strategies, achieving the best performance when the synthesized projection quantity matches the original projection number (ratio = 1). This demonstrates that the proposed adaptive selection mechanism effectively balances angular diversity and data redundancy, improving both convergence and projection refinement quality.

\begin{table}[htbp]
  \centering
  \caption{Ablation results of different projection selection strategies under varying synthesized projection ratios (average over AAPM L067 and L096).}
  \label{tab:apgps}
  \footnotesize
  \resizebox{\columnwidth}{!}{
  \begin{tabular}{c c c c c}
    \toprule
    \diagbox{Strategy}{Proj Ratio} & 1/4 & 1/2 & 1 & 2 \\
    \midrule
    Central & 37.37/0.9649 & 37.55/0.9661 & 37.54/0.9669 & 36.88/0.9615 \\
    Random  & 36.94/0.9628 & 37.02/0.9617 & 37.37/0.9628 & 36.91/0.9632 \\
    APGPS   & 37.39/0.9651 & 37.91/0.9724 & \textbf{38.17/0.9807} & 37.09/0.9634 \\
    \bottomrule
  \end{tabular}
  }
\end{table}

In addition, the right figure in Fig.~\ref{fig:ablation_bar}(b) presents a bar chart comparison of Central, Random, and APGPS at the 1 projection ratio. The bars show higher PSNR/SSIM for APGPS, indicating that adaptive, geometry-aware selection is more effective than fixed or random angle selection under the iterative framework.

\textbf{Data Reuse Strategy:} This study investigates the data reuse strategy within the DRPR module of the proposed iterative framework. A key challenge arises because the DR projections used to train NAF and the refined DR projections reintroduced during iteration may have inconsistent data distributions. To address this, we introduce a DRAT that ensures consistency between input and output data during projection refinement.

We compare three strategies: (1) No Processing, where refined projections are directly reused without adjustment; (2) Normalization / Inverse Normalization, where data is linearly scaled to a fixed range and then restored; and (3) DRAT, which applies a prior-guided affine transformation dynamically adapted to the statistical properties of each projection. As shown in Table~\ref{tab:drat}, DRAT achieves the best performance in both DR image refinement and 3D reconstruction. Direct reuse leads to performance degradation due to distribution mismatch between synthesized and real projections, while normalization partially alleviates this but introduces transformation errors. DRAT maintains stable data consistency and minimizes reconstruction error.

\begin{table}[htbp]
  \centering
  \caption{Ablation of data reuse strategies in the DRPR module.}
  \label{tab:drat}
  \footnotesize
  \begin{tabular}{c c c}
    \toprule
    Reuse Strategy & L067 & L096 \\
    \midrule
    None & 34.91/0.9640 & 37.83/0.9714 \\
    Norm/Inv-Norm & 35.33/0.9657 & 38.09/0.9793 \\
    DRAT & \textbf{37.21/0.9780} & \textbf{39.13/0.9833} \\
    \bottomrule
  \end{tabular}
\end{table}

\begin{figure}[htbp]
    \centering
    \includegraphics[width=\columnwidth]{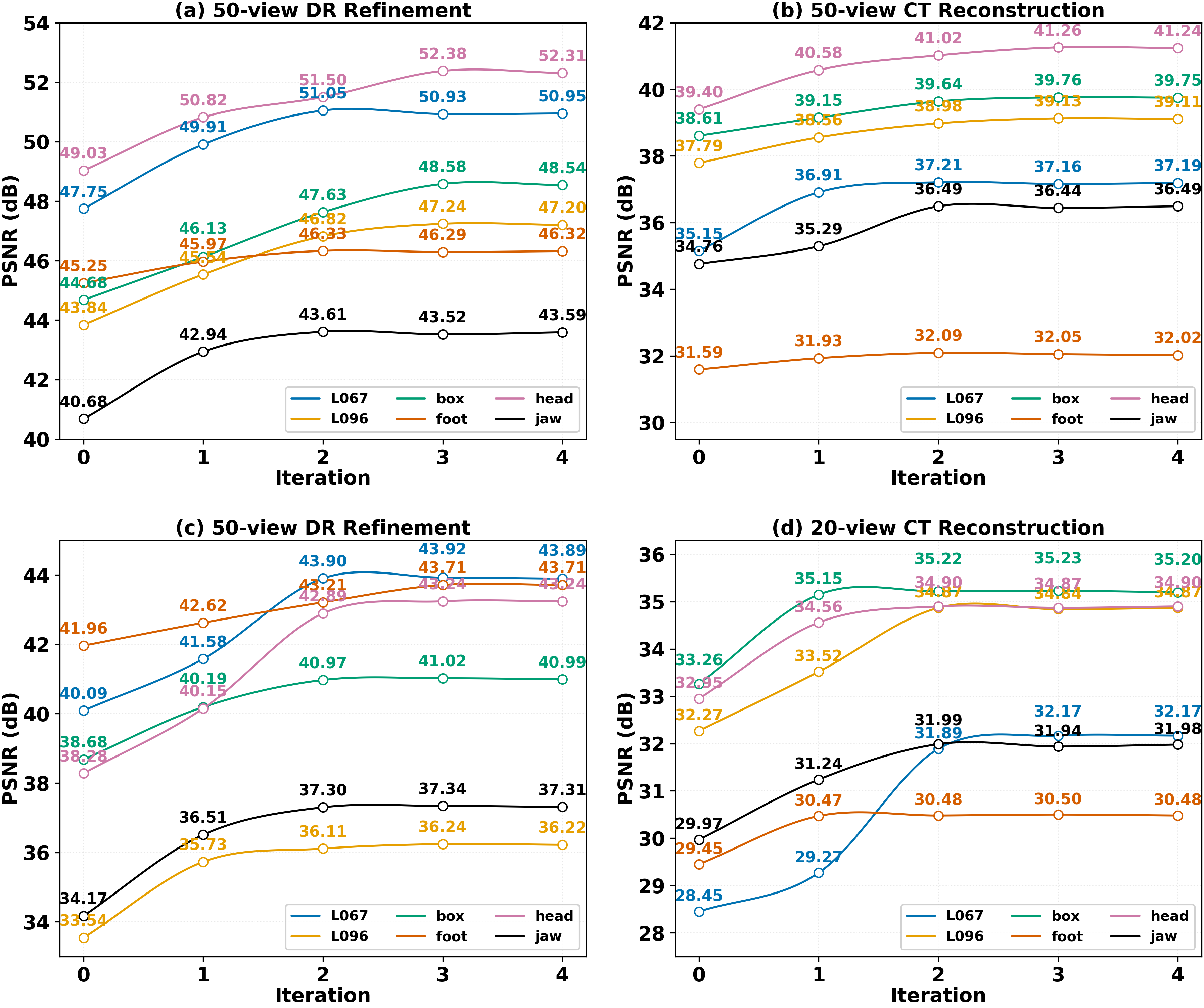}
    \caption{Relationship between the number of outer iterations and quantitative metrics (PSNR/SSIM) for four tasks:(a) 50-view DR Refinement, (b) 50-view CT Reconstruction, (c) 20-view DR Refinement, (d) 20-view CT Reconstruction.}
    \label{fig:iter_count}
\end{figure}

\textbf{Number of Iterations:} We study how the number of outer iterations affects performance in the proposed iterative process. As shown in Fig.~\ref{fig:iter_count}, PSNR/SSIM improve rapidly within the first two to three iterations and then plateau with minor fluctuations. This behavior arises because, as training proceeds, newly incorporated views contribute diminishing additional information, while the quality of synthesized-and-corrected projections increases, reducing the marginal benefit of further updates. A practical balance between reconstruction quality and computation is thus achieved with 2--4 iterations.

\section{CONCLUSION AND DISCUSSION}\label{sec:conclusion}
This study addresses the challenging problem of ultra-sparse-view imaging in multi-source stationary CT, where severe underdetermination often leads to incomplete and artifact-prone reconstructions. Traditional model-based or learning-based methods struggle in this regime due to insufficient projection constraints. To overcome this limitation, this work proposes Diff-NAF, an iterative framework that unifies physical modeling and generative priors for progressive reconstruction enhancement.

Unlike prior approaches that treat neural fields and diffusion models independently, Diff-NAF couples them through a DRPR module. This integration injects learned distribution priors into the physical forward model, compensating for missing information under ultra-sparse-view sampling. In parallel, the APGPS strategy adaptively selects novel projection views to improve angular coverage and convergence efficiency. Through iterative training, refined projections are reused as pseudo-labels, continuously improving reconstruction fidelity and structural completeness.

Experiments on multiple simulated and real datasets show consistent PSNR/SSIM gains over classical reconstruction algorithms and state-of-the-art implicit neural representations. Diff-NAF also exhibits robust generalization across different anatomical and object structures.

Beyond quantitative gains, the results highlight the synergy between physics-based models and generative priors. NAF enforces geometric and data consistency, while the diffusion prior enhances global coherence and regularizes the solution space. Their interaction yields a closed-loop process that progressively improves both data fidelity and structural realism.

Although the current framework focuses on refinement in the DR domain and achieves strong results, minor inter-view inconsistencies may remain, which can affect the final volumetric reconstruction. A plausible reason is that the diffusion prior can partially disrupt consistency across angles, which introduces discontinuities during training.

Future work will extend the framework with joint refinement in both the DR and sinogram projection domains and with explicit physics-based constraints across slices and across angles, aiming to improve inter-angle and inter-layer consistency and to further enhance stability under ultra-sparse-view conditions.

\bibliographystyle{IEEEtran}
\bibliography{references}  

\end{document}